\makeatletter
\let\@twosidetrue\@twosidefalse
\let\@mparswitchtrue\@mparswitchfalse
\makeatother

\documentclass{llncsown}
\usepackage{graphicx}
\usepackage{multicol}

\usepackage[utf8]{inputenc}

\usepackage{lineno}
\usepackage{times} 

\usepackage[small,it]{caption}
\usepackage{wrapfig}

\newcommand{\squishlist}{
 \begin{list}{$-$}
  { \setlength{\itemsep}{0pt}
     \setlength{\parsep}{3pt}
     \setlength{\topsep}{3pt}
     \setlength{\partopsep}{0pt}
     \setlength{\leftmargin}{1.5em}
     \setlength{\labelwidth}{1em}
     \setlength{\labelsep}{0.5em} } }

\newcommand{\squishlisttwo}{
 \begin{list}{$-$}
  { \setlength{\itemsep}{0pt}
    \setlength{\parsep}{0pt}
    \setlength{	opsep}{0pt}
    \setlength{\partopsep}{0pt}
    \setlength{\leftmargin}{2em}
    \setlength{\labelwidth}{1.5em}
    \setlength{\labelsep}{0.5em} } }

\newcommand{\squishend}{
  \end{list}  }
  
\usepackage[spanish]{babel}
\renewcommand{\tablename}{Tabla}
\renewcommand{\refname}{Referencias}

\begin{document}
\title{Estimaci\'on de carga muscular mediante im\'agenes}
  %If you're using runningheads you can add an abreviated title for the running head on odd pages using the following
%\titlerunning{abreviated title goes here}
%and an alternative title for the table of contents:
%\toctitle{table of contents title}

%\subtitle{Subtitle Goes Here}

%For a single author
%\author{Author Name}

%For multiple authors:
\author{Leandro Abraham \inst{1} \inst{2} \inst{3} , Facundo Bromberg \inst{1} \inst{3} y Raymundo Forradellas \inst{2}}
% \author{}

%If using runnningheads you can abbreviate the author name on even pages:
%\authorrunning{abbreviated author name}
%and you can change the author name in the table of contents
%\tocauthor{enhanced author name}

%For a single institute
%\institute{Institute Name \email{email address}}

% If authors are from different institutes 
% \institute{}

\institute{Laboratorio DHARMa, DISI, Universidad Tecnol\'ogica Nacional, FRM
\email{leandro.abraham@frm.utn.edu.ar y fbromberg@frm.utn.edu.ar} 
\and CEAL, Universidad Nacional de Cuyo, Facultad de Ingenier\'ia \email{ceal@fing.uncu.edu.ar} 
\and  Consejo Nacional de Investigaciones Cient\'ificas y T\'ecnicas (CONICET) }

%to remove your email just remove '\email{email address}'
% you can also remove the thanks footnote by removing '\thanks{Thank you to...}'

\maketitle
\vspace{-0.2in}
\begin{abstract}
% A problem of great interest in disciplines like occupational, ergonomics and sports (among others)
% is the measurement of biomechanical variables involved in human movement (as internal muscle forces and joint torques).
% Nowadays this problem is solved by a process with two steps. First capturing data with impractical, intrusive and expensive devices.
% Then this data is used as input in complex models to obtain the variables as outputs.
% The work we present in this papaer represents the first steps towards capturing some of the models inputs in an automated, non-intrusive and economic way.
% Later the idea is to automate these variables (internal forces and torques of joints) complex calculation process.
% We present a method for estimating the discrete level of muscular load that the muscles of the arm are exercising.
% This method uses as input static images.
% In order to classify the load level of the image our approach uses various visual feature extraction techniques
% (Bag of Keypoints, Local Binary Patterns, Color Histograms, Contour Moments) combined with a supervised machine learning technique (SVM for classification).
% In the best cases we obtained measures of performance (Accuracy, Precision, Recall, FMeasure, and Specificity) above 90\%.

% % 
% \textbf{Resumen.}
Un problema de gran inter\'es en disciplinas como  la ocupacional, ergon\'omica y deportiva, 
es la medici\'on de variables biomec\'anicas  involucradas en el movimiento humano (como las fuerzas musculares internas y torque de articulaciones).
Actualmente este problema se resuelve en un proceso de dos pasos.
Primero capturando datos con dispositivos poco pr\'acticos, intrusivos y costosos.
Luego estos datos son usados como entrada en modelos complejos para obtener las variables biomec\'anicas como salida. 
El presente trabajo representa una alternativa automatizada, no intrusiva y econ\'omica al primer paso, proponiendo la captura de estos datos a trav\'es de im\'agenes.
% El trabajo que presentamos en este paper representa los primeros pasos hacia la captura de algunas de las entradas de los modelos 
% de manera automatizada, no intrusiva y econ\'omica.
En trabajos futuros la idea es automatizar todo el proceso de c\'alculo de esas variables.  
% (fuerzas internas y torques de articulaciones). 
En este trabajo elegimos un caso particular de medición de variables biomecánicas: el problema de estimar el nivel discreto de carga muscular 
que est\'an ejerciendo los m\'usculos de un brazo.
% Para estimar a partir de im\'agenes, el nivel discreto de carga
% presentamos un procedimiento para estimar el nivel discreto de carga muscular que est\'an ejerciendo los m\'usculos de un brazo. 
% Esto lo resolvemos usando como entrada im\'agenes est\'aticas.
% Para clasificar el nivel de carga de la imagen 
Para estimar a partir de im\'agenes est\'aticas del brazo ejerciendo la fuerza de sostener la carga, 
el nivel de la misma, realizamos un proceso de clasificaci\'on.
Nuestro enfoque utiliza Support Vector Machines 
% (una t\'ecnica de aprendizaje de m\'aquinas supervisado) 
para clasificaci\'on, 
combinada con una etapa de pre-procesamiento que extrae caracter\'isticas visuales utilizando variadas t\'ecnicas 
(Bag of Keypoints, Local Binary Patterns, Histogramas de Co\-lor, Momentos de Contornos)
% Nuestro enfoque utiliza distintas t\'ecnicas de extracci\'on de caracter\'isticas visuales 
% (Bag of Keypoints, Local Binary Patterns, Histogramas de Co\-lor, Momentos de Contornos) combinados con una t\'ecnica de clasificaci\'on
% (Support Vector Machine). 
En los mejores casos (Local Binary Patterns y Momentos de Contornos) obtenemos medidas de performance en la clasificaci\'on (Precision, Recall, F-Measure
% , Specificity
y Accuracy) superiores al 90\%.
\end{abstract}

 \begin{keywords}
  biomechanical variables, muscle arm load, support vector machine, local binary patterns, bag of key points, contour moments, color histograms  
 \end{keywords}

\vspace{-0.1in}

\section{Introducci\'on}\label{sec:intr}
\vspace{-0.15in}
La biomec\'anica es una disciplina cient\'ifica que tiene por objeto el estudio de las estructuras de car\'acter mec\'anico, modelos, fen\'omenos y leyes que sean 
relevantes al movimiento y al equilibrio de los seres vivos; fundamentalmente del cuerpo humano. Las variables biomec\'anicas  
% que m\'as interesan 
m\'as estudiadas
al momento de analizar 
el movimiento humano son: las fuerzas internas ejercidas por los m\'usculos involucradas en los mismos y los torques que se producen en las articulaciones al ejecutarse 
estos movimientos. 
El an\'alisis de las variables biomec\'anicas permite identificar: 
movimientos perjudiciales para la salud, situaciones de sobre esfuerzo, posiciones in/adecuadas, movimiento \'optimo, entre otras situaciones.
Esto conlleva a que tenga gran impacto en disciplinas como la ocupacional \cite{chaffin1984occupational},  
% cuyo objetivo es mejorar las condiciones laborales de los empleados y prevenir posibles daños en el cuerpo y trastornos de salud debido a malos movimientos o posturas en tareas laborales. 
la ergon\'omica \cite{kumar1999biomechanics} y 
el deporte  \cite{mcginnis2013biomechanics}.
% hay1993biomechanics,

Para 
% estudiar la biomec\'anica del movimiento humano y 
encontrar las fuerzas musculares y torques de articulaciones actuantes 
en un movimiento, com\'unmente se aplican modelos din\'amicos complejos \cite{manal2004biomechanics,pandy2004biomechanics,shao2004electromyography}.
% (secci\'on \ref{sec:mod}).
La informaci\'on de entrada para estos modelos es obtenida mediante dispositivos como:
electromiografos (EMG) que miden nivel de activación muscular de forma el\'ectrica 
(requiere adhesi\'on de electrodos cableados al cuerpo o introducci\'on de agujas en los m\'usculos),
goni\'ometros que miden \'angulos de las partes del cuerpo (implica adherir sensores y elementos al cuerpo de las personas), 
sistemas de captura de movimiento para medir posiciones y \'angulos de las partes del cuerpo (requiere adherir marcadores y montar sistemas multic\'amaras), 
entre otros.
% Como comentamos, 
Los dispositivos utilizados para capturar la informaci\'on necesaria en estos modelos son costosos e intrusivos. 
Adem\'as no son aptos para utilizarlos fuera del entorno de un laboratorio 
ya que requieren el montaje de dispositivos especiales (cableados) sobre el cuerpo de las personas (electrodos o agujas en el cuerpo, traje con marcadores, goni\'ometros) 
y en el entorno (sistema de m\'ultiples c\'amaras).
  
% Teniendo en cuenta la complejidad en la adquisici\'on de datos de entrada y de lo modelos presentados
% La adquisici\'on de los datos para estos modelos as\'i tambi\'en como su c\'alculo, son procesos complejos.
En este trabajo proponemos el punto de partida de una investigaci\'on cuyo objetivo final consiste en poder automatizar 
lo m\'as posible la medici\'on no intrusiva de las variables biomec\'anicas de inter\'es 
% (fuerzas musculares internas y/o torques de articulaciones) 
involucradas en un movimiento 
humano, y de las variables intermedias que necesiten ser usadas en el proceso.
% (EMG). 
Como primer paso, proponemos en este trabajo una soluci\'on al problema de 
estimar el nivel discreto de carga muscular que se encuentra realizando un brazo humano levantando un objeto de peso conocido, solamente utilizando informaci\'on 
visual extra\'ida desde fotograf\'ias del brazo. Tanto el problema particular, como la l\'inea de investigaci\'on a la que pertenece 
% como as\'i 
y
tambi\'en 
la soluci\'on que se propone, 
% son novedosos 
mejoran y simplifican a las t\'ecnicas existentes
ya que consisten en un enfoque autom\'atico y no intrusivo.
% para estimar carga muscular y variables biomec\'anicas. 
% Este enfoque es novedoso ya que 
Hasta donde los 
autores conocemos, no se han presentado soluciones que estimen nivel de carga o variables biomec\'anicas s\'olo a partir de informaci\'on de im\'agenes 
como se podrá observar en la Secci\'on \ref{sec:trel} lo que hace novedoso nuestro enfoque.

Con el objetivo de resolver el problema planteado es que presentamos en este trabajo un procedimiento que permite estimar el nivel de peso discreto 
levantado por el brazo de una persona a partir de una imagen en la Secci\'on \ref{sec:ecmmi}. 
Este procedimiento consiste en 
utilizar una t\'ecnica de aprendizaje de m\'aquinas supervisado para clasificar las im\'agenes, 
combinada con una etapa de pre-procesamiento que caracteriza las mismas utilizando variadas t\'ecnicas de extracci\'on de caracter\'isticas visuales. 
% la aplicaci\'on de  t\'ecnicas de extracci\'on de caracter\'isticas visuales a im\'agenes est\'aticas y aprendizaje de m\'aquinas para clasificar  
% cu\'al de 3 niveles discretos de carga se encuentra ejerciendo la persona en la imagen. 
Posteriormente en la Secci\'on \ref{sec:exp} presentamos 
experimentaci\'on y resultados que demuestran de forma inicial el buen funcionamiento del procedimiento propuesto. 
Con el motivo de encuadrar este trabajo en el estado del arte y dimensionar su impacto es que en la Secci\'on \ref{sec:trel} presentamos
un breve repaso y discusi\'on sobre los trabajos m\'as relacionados al problema y al tema de investigaci\'on en general.
Finalmente presentamos nuestras conclusiones y nuevos lineamientos para seguir trabajando en la Secci\'on \ref{sec:concl}.

\vspace{-0.1in}
% \section{Estimaci\'on del nivel de carga muscular mediante im\'agenes}\label{sec:ecmmi}
\section{Enfoque}\label{sec:ecmmi}
\vspace{-0.15in}

% Con el objetivo de solucionar el problema propuesto en la Secci\'on \ref{sec:intr}, diseñamos e implementamos un procedimiento que explicaremos en esta secci\'on.
% \vspace{-0.2in}

%   \subsection{Captura de las im\'agenes }
  \subsection{Captura de las im\'agenes y Preprocesamiento de los datos }
  \vspace{-0.1in}
  Se procede de la siguiente manera para capturar las im\'agenes: 
  1) el sujeto de prueba sostiene una carga de peso $Pi$ con el brazo estirado, 
  a piel descubierta; 
  2) se toma una fotograf\'ia solamente del brazo (sin capturar el objeto) con un fondo azul para facilitar posteriormente el proceso de segmentaci\'on 
  del brazo; 
  3) para un $P_i$ dado se repiten los pasos 1 y 2, N veces para obtener N ejemplos del ese nivel de carga; 
  4) se repite el proceso anterior con el resto de los pesos.
  
%   \setlength{\columnsep}{0pt}%
%   \begin{wrapfigure}{5pt}{0.45\textwidth}
% \vspace{-0.2in}
%   \centering
%   \includegraphics[scale=0.025]{segm.eps}  
%   \caption{\small \sl Segmentaci\'on de imagen\label{fig:segm}}  
%   \vspace{-0.2in}
% \end{wrapfigure}
  
  Luego de capturar las im\'agenes es necesario un realizar la segmentaci\'on del brazo.
  Para ello se propone un proceso semiautom\'atico de segmentaci\'on 
  que consiste en identificar en la imagen los p\'ixeles que son del color del fondo (azul) y eliminarlo como background. 
  La segmentaci\'on se realiz\'o aplicando el algoritmo de clustering  K-Means \cite{chitade2010colour} tomando como datapoints los valores RGB de cada p\'ixel 
  de las im\'agenes. 
  Posteriormente se
%   la transform\'o a escala de grises para poder hacer 
aplic\'o
  una operaci\'on morfol\'ogica de erosi\'on para remover posible ruido. 
  Finalmente a los p\'ixeles del background en la imagen original se les asign\'o el valor cero. 

  \vspace{-0.2in}
  \subsection{Extracci\'on de caracter\'isticas visuales}\label{sec:ecv}
  \vspace{-0.15in}
  El pr\'oximo paso es extraer un vector de caracter\'isticas de las im\'agenes para usarlo como entrada
  en el proceso de aprendizaje. Las t\'ecnicas que presentaremos en esta secci\'on se usaron para tal fin y se programaron usando las versiones de los 
  algoritmos implementadas en la librer\'ia OpenCV \cite{opencv_library}. Como resultado de esta etapa se obtuvieron un dataset por cada t\'ecnica de extracci\'on de 
  caracter\'isticas implementada. Los vectores de caracter\'isticas de cada imagen se etiquetaron con el tipo de objeto levantado, lo que tiene una relaci\'on directa con 
  el nivel de carga, ya que cada tipo de objeto tiene un peso distinto.
  De esta forma cada dataset contiene un datapoint por imagen que consiste en una tupla formada por el vector de caracter\'isticas extraídas para la imagen, 
  y la clasificaci\'on correcta de la imagen.

%   En este paso se eval\'ua y experimenta el uso de alternativas maduras y com\'unmente utilizadas.
  \vspace{-0.1in}
    \subsubsection{Bag of Keypoints (BKP)}
    \vspace{-0.1in}
    Un enfoque para extraer caracter\'isticas visuales muy utilizado en categorizaci\'on visual es el de BKP
    \cite{csurka2004visual}. 
%     ,jurie2005creating,nowak2006sampling
     Con este enfoque primero se encuentran porciones de la imagen que presentan caracter\'isticas 
     que pueden ser igualmente detectadas bajo variaci\'on de escala, iluminaci\'on o ruido (puntos de inter\'es).
     Generalmente se caracterizan por ser zonas de gran contraste en la imagen. En nuestro trabajo utilizamos el algoritmo 
     SURF \cite{bay2006surf} para encontrar estos puntos de inter\'es. Cada uno se caracteriza 
     mediante un vector que contiene informaci\'on del mismo (posici\'on, orientaci\'on, entre otras).
     Posteriormente se entrena una t\'ecnica de clustering para agrupar los puntos de inter\'es 
%      de las im\'agenes 
     en un diccionario de 
     porciones de im\'agenes seg\'un sus vectores de caracter\'isticas, en una cantidad definida de clusters. 
     En nuestro caso el tamaño del diccionario es de 800 grupos y se obtuvo aplicando la t\'ecnica de clusterizaci\'on KMeans a los vectores descriptivos de los 
     puntos de inter\'es.
     Luego se analizan todas las im\'agenes para obtener un histograma de aparici\'on de cada punto de inter\'es en cada imagen nueva. 
     Este histograma se calcula para cada imagen sobre los distintos clusters, por lo que tendr\'a 800 intervalos para nuestro caso. 
     Finalmente se devuelven estos histogramas como vectores de caracter\'isticas. En nuestro caso entonces la cantidad de variables del vector de caracter\'isticas 
     por imagen es de $n=800$.

     \vspace{-0.1in}
    \subsubsection{Local Binary Patterns (LBP)}
    \vspace{-0.1in}
    Otra forma de extraer caracter\'isticas de im\'agenes muy utilizada es LBP \cite{ojala1996comparative}, 
    que extrae informaci\'on de textura de la imagen, y ha sido muy utilizada para caracterizaci\'on de la piel en el \'area de  reconocimiento de expresiones faciales 
    \cite{zhao2007dynamic}. Este enfoque es invariante a rotaciones y robusto a variaciones en escala de grises.
    
    El proceso 
%     a grandes rasgos 
    consiste en los siguientes pasos: 1) dividir cada imagen en una grilla de tamaño $NxN$, obteniendo as\'i $NxN$ regiones de inter\'es (ROI);
    2) generar un c\'odigo binario de 8 bits para cada p\'ixel en una ROI comparando el p\'ixel con sus 8 p\'ixeles vecinos (superior, inferior, 
     derecho, izquierdo y diagonales) y decidiendo por el valor 0 si el p\'ixel central es mayor a su vecino y 1 en caso contrario; 
    3) estos c\'odigos posteriormente son transformados a un valor decimal y cuantizados en un histograma por ROI (que tiene 59 intervalos en nuestro caso);
    4) los histogramas de cada ROI se concatenan en un nuevo vector que representa la imagen completa dando como resultado un vector de caracter\'isticas 
     de tamaño $n=N*N*59$. En nuestro caso particionamos la imagen en $3x3$ por lo que el vector tiene un tamaño $n=3*3*59=531$.

%     %     \begin{enumerate}
%     \squishlist
%      \item Dividir cada imagen en una grilla de tamaño $NxN$, obteniendo as\'i $NxN$ regiones de inter\'es.
%      \item Generar un c\'odigo binario para cada p\'ixel en una regi\'on de inter\'es: Este c\'odigo se obtiene comparando el p\'ixel con sus 8 vecinos (superior, inferior, 
%      derecho, izquierdo y diagonales) y decidiendo por el valor 1 si el p\'ixel central es mayor a su vecino y 0 en caso contrario. 
%      \item Cuantizar los c\'odigos: Estos c\'odigos posteriormente son cuantizados en un histograma que tiene 59 clases en nuestro caso. 
%      \item Los histogramas de cada regi\'on se concatenan en un nuevo vector que representa la imagen completa. Esto nos d\'a como resultado un vector de caracter\'isticas 
%      de tamaño $N=3*3*59=531$.
%      \squishend
% %     \end{enumerate}

%     El proceso descrito anteriormente lo implementamos mediante un programa en C++, usando OpenCV (\cite{opencv_library}).
     \vspace{-0.1in}
    \subsubsection{Histograma de color (HC)}
    \vspace{-0.1in}
    Otra t\'ecnica utilizada com\'unmente en caracterizaci\'on de im\'agenes  para clasificaci\'on \cite{chapelle1999support} y para recuperaci\'on de im\'agenes
    \cite{sural2002segmentation}, son los histogramas de color. Esta t\'ecnica consiste en cuantizar los valores de los 
    p\'ixeles de una ROI en histogramas de tamaño definido.

%     En nuestro caso implementamos un programa (usando OpenCV (\cite{opencv_library})) para calcular 
  En nuestro caso calculamos los
    histogramas de color de los dos primeros (HS) canales del espacio HSV 
    (Hue Saturation Value \cite{smith1978color}) de una imagen particionada en celdas. 
    Para nuestros experimentos la imagen se particiona en una grilla de $3x3$ celdas o ROI y se usan 
    5 intervalos para cada canal del histograma. 
    De esta forma el valor de cada p\'ixel de cada canal aporta a la suma de uno de los 5 intervalos del histograma del canal seg\'un corresponda
    Est\'a t\'ecnica genera un vector de caracter\'isticas de tamaño $n=3*3*5*5=225$.
    El hecho de trabajar sobre el espacio de color HSV provee a la t\'ecnica robustez a los cambios en iluminaci\'on.
 \vspace{-0.1in}
 
    \subsubsection{Momentos de contornos (MC)}
    \vspace{-0.1in}
    Otra forma de caracterizar una imagen es a trav\'es de informaci\'on de su contorno. Una de las formas m\'as simples de comparar contornos es a 
    trav\'es de sus momentos. 
    El momento de un contorno es un promedio pesado de las intensidades de los p\'ixeles.
    Definimos el momento $m_{pq}$ de un contorno como : 
    \vspace{-0.1in}
    \[
    m_{p,q} = \sum_{i=1}^{n}I(x,y) x^p y^q
    \]
%     Donde $p$ es el orden $x$ y $q$ es el orden $y$. 
    A $p$ y $q$ se los denomina \'ordenes del momento.
    La sumatoria es sobre todos los p\'ixeles del contorno ($n$). $I(x,y)$ es el valor de intensidad del p\'ixel (x,y) de la imagen.
    Los contornos de la imagen se obtienen con un algoritmo de detecci\'on de bordes.	
    
    Es posible definir distintos tipos de momentos. Los momentos espaciales son 10 y se encuentran variando los \'ordenes de $x$ e $y$ ($p$ y $q$) en la fórmula anterior. 
    Los momentos centrales son 7 y se obtienen agregando informaci\'on de los momentos de orden $0$ y $1$ en la formula original. 
    Los momentos normalizados son 7 y se estiman a partir de los momentos centrales divididos por una potencia del momento $m_{00}$. 
    Finalmente los Hu Invariant moments  \cite{hu1962visual}  son 7 y se estiman como combinaciones lineales de los momentos centrales. 
    Estos últimos son muy \'utiles ya que permiten caracterizar el contorno de forma invariante a escala, rotaci\'on y reflexi\'on.
    Concatenando todos estos momentos generamos un vector de $n=10+7+7+7=31$ caracter\'isticas. 
    
%     A partir de aqu\'i es posible definir distintos tipos de momentos:
%     \begin{itemize}
%      \item Momentos espaciales: son 10 y se encuentran variando los \'ordenes de x e y (p y q) en la f\'ormula anteriores
%      \item Los momentos centrales: son 7 y se estiman mediante la f\'ormula: 
%       \[
%       \mu_{p,q} = \sum_{i=1}^{n}I(x,y) (x - x_{avg})^p (y-u_{avg})^q
%       \]
%      \item Los momentos normalizados : son 7 y se estiman con la f\'ormula:
%       \[
%       \eta_{p,q} = \frac{\mu_{p,q}}{m_{00}^{(p+q)/2+1}}
%       \]
%       \item Los Hu Invariant moments  (\cite{hu1962visual}) : Son 7 y se estiman como combinaciones lineales de los momentos centrales
%     \end{itemize}
%     
    
%       En nuestro trabajo implementamos un programa (con OpenCV(\cite{opencv_library})) que realiza los siguientes pasos: 1) Detecta contornos usando el algoritmo de Canny
%       (\cite{canny1986computational}), 2)Encuentra los momentos espaciales, centrales y normalizados de los bordes de la imagen, 3) Calcula los Hu Invariant Moments a 
%       partir de los momentos centrales y 4) Concatena toda esa informaci\'on en un vector por cada imagen. 
%       Este programa genera un vector de $N=20+7+7+7=31$ caracter\'isticas.
   
   \vspace{-0.2in}
%   \subsection{Proceso de aprendizaje y estimaci\'on de carga para ejemplos nuevos}
  \subsection{Procedimiento de aprendizaje}
  \vspace{-0.15in}
  En este paso se aprende el modelo subyacente en los datasets de entrenamiento mediante la aplicaci\'on de una t\'ecnica de aprendizaje de m\'aquinas 
  supervisado de manera de poder clasificar 
  el nivel de carga muscular de nuevos ejemplos.   
%   Como en todo proceso de aprendizaje el dataset completo se dividi\'o 
%   en una proporci\'on para entrenamiento y la proporci\'on restante para testeo.
  Nosotros utilizamos la t\'ecnica de Support Vector Machine (SVM) \cite{cortes1995support,bishop2009pattern} debido a que 
  es robusta y muy usada para tareas de clasificaci\'on.
  Utilizamos la funci\'on correspondiente (svm) implementada en el paquete \textit{e1071} del lenguaje de programaci\'on R \cite{meyer2012support},
  con los valores por defecto para 
  sus par\'ametros (kernel de tipo radial con gamma=0.0018). 
%   Usando estos recursos aprendimos el modelo de los datos subyacentes sobre el dataset de entrenamiento. 
  Debido a que se obtuvieron 4 datasets distintos (resultado del proceso de la Secci\'on \ref{sec:ecv}), el proceso de entrenamiento y 
  clasificaci\'on se realiz\'o 4 veces.
  Finalmente, usando la misma t\'ecnica de aprendizaje predecimos los valores o niveles de carga realizando clasificaci\'on tri-clase sobre cada dataset de testeo a partir 
  del modelo aprendido (consideramos im\'agenes levantando solo tres cargas diferentes)
%   0Kg, 2.25Kg y 5Kg).
\vspace{-0.15in}
\section{Experimentaci\'on y resultados}\label{sec:exp}
\vspace{-0.15in}
Realizamos el proceso de aprendizaje y clasificaci\'on sobre un conjunto de 92 im\'agenes segmentadas, de brazos sosteniendo 3 pesos distintos.
Los pesos y objetos utilizados son: el objeto O0 que pesa 0Kgs, el objeto O5 que pesa 2.25Kgs y el objeto O6 que pesa 5Kgs. 
  Estos pesos corresponden con las clases 0, 5 y 6 respectivamente.
Se tienen 30 instancias de la clase 0, 33 instancias de la clase 5 y 29 instancias de la clase 6. Cada dataset se fragment\'o en 
un conjunto de entrenamiento conformado por el 70\% del dataset original 
% (64 observaciones)
y el conjunto de testeo conformado por el 30\% restante 
% (28 observaciones) 
tomadas al azar del dataset completo. En distintas instancias del experimento se entren\'o el clasificador usando datos extra\'idos con los 
4 tipos de caracter\'isticas
presentadas, se reportaron las medidas de rendimiento que presentaremos en breve y se realiz\'o un an\'alisis comparativo de las mismas.

%  Siendo $TP=True Positives$ , $TN=True Negatives$ , $FP=False Positives$ y $FN=False Negatives$; las medidas que presentaremos son:
%  $Accuracy=(TP+TN)/(TP+TN+FP+FN)$, $Precision=(TP)/(TP+FP)$, $Recall=(TP)/(TP+FN)$, $FMeasure=2((Precision  * Recall)/(Precision + Recall))$ y 
%  $Specificity=(TN)/(TN+FP)$

Como medidas de rendimiento se consideraron \textit{Precision} (P), \textit{Recall} (R), \textit{F-Measure} (FM)
% , \textit{Specificity} (\hat{S}) 
y \textit{Accuracy} (A), para el caso multi-clase para cada experimento.
Para la i-esima clase, la $P_i$ corresponde a la cantidad de ejemplos de la i-esima clase clasificados correctamente, 
dividido por la cantidad de ejemplos de otras clases clasificados incorrectamente como pertenecientes a la i-esima clase. 
En t\'erminos de la matriz de confusi\'on, estas cantidades corresponden a la cantidad reportada en la diagonal del i-esimo rengl\'on y 
la suma de las cantidades del i-esimo rengl\'on, respectivamente. 
El $R_i$ de la i-esima clase es similar, solo que el denominador corresponde al total de ejemplos de la clase clasificados incorrectamente como pertenecientes a alguna 
otra clase, i.e., la suma de las cantidades de la i-esima columna. 
El c\'alculo de la $FM_i$ se computa de igual manera que el caso binario pero usando los valores $P_i$ y $R_i$ , 
i.e., $FM_i=2*((P_i*R_i)/(P_i+R_i))$. 
% NOTA 2 A LEANDRO:  Seguir el mismo patron para explicar specificity  (que no termino de entenderlo tal como está redactado)
Por \'ultimo, la $A_i$, que corresponde a la fracción de ejemplos clasificados correctamente como pertenecientes a la clase i-esima, 
se computa como la suma de las cantidades de la diagonal 
(i.e., clasificados correctamente como pertenecientes a la clase) mas la suma de las cantidades no pertenecientes ni al i-esimo renglón ni a la i-esima columna 
(i.e., clasificados correctamente como no pertenecientes a la i-esima clase), 
dividida  por la suma de todas las cantidades de la matriz (i.e., el número total de ejemplos).

Adem\'as  para cada experimento es calculado el promedio de estas medidas ($\hat{P}$, $\hat{R}$, $\hat{FM}$
% , $Mean\hat{S}$
y $\hat{A}$) sobre todas las clases, junto con la Overall Accuracy (OvA). Esta \'ultima medida es la accuracy sobre toda la matriz de confusi\'on y 
se calcula sumando los elementos de la diagonal de la matriz dividido por la suma de todos los elementos de la misma).

  Resumimos en el Cuadro \ref{tab:gral} los valores promedio y de OvA obtenidos para cada medida de rendimiento sobre cada dataset obtenido con las distintas
  t\'ecnicas de extracci\'on de caracter\'isticas.
%   y presentaremos un breve an\'alisis.

% %  \resizebox{20}{30pt}{
%    \begin{wrapfigure}{r}{0.6\textwidth}
% %   \begin{figure}
%   \begin{center}  
%   \includegraphics[scale=0.030]{segm.eps}  
%   \caption{\small \sl Semgentaci\'on de imagen\label{fig:segm}}  
%   \end{center}  
% %   \end{figure}  
% %  }
% \end{wrapfigure}

%Sin Specificity
\begin{table}[htbp]
% \begin{table}[H]
% \centering
\begin{center}
\resizebox{!}{30pt}{
\begin{tabular}{|c|c|c|c|c|c|c|}
\hline
\textbf{Exp.} & \textbf{OvA} & \textbf{$\hat{A}$} & \textbf{$\hat{P}$} & \textbf{$\hat{R}$} & \textbf{$\hat{FM}$}  \\ \hline
\textbf{BKP} & 82.14\% & 88.09\% & 80.83\% & 82.78\% & 80.83\% \\ \hline
\textbf{LBP} & \textbf{100\%} & \textbf{100\%} & \textbf{100\%} & \textbf{100\%} & \textbf{100\%}  \\ \hline
\textbf{HC} & 75\% & 83.33\% & 76.06\% & 73.56\% & 73.29\%  \\ \hline
\textbf{MC} & 92.85\% & 95.23\% & 93.93\% & 93.93\% & 93.93\%  \\ \hline
\end{tabular}
}
\end{center}
\caption{Resultados generales}
\label{tab:gral}
\end{table}

% %Con Specificity
% \begin{table}[htbp]
% % \begin{table}[H]
% % \centering
% \begin{center}
% \resizebox{!}{30pt}{
% \begin{tabular}{|c|c|c|c|c|c|c|}
% \hline
% \textbf{Exp.} & \textbf{Ov\hat{A}} & \textbf{Mean\hat{A}} & \textbf{Mian\hat{P}} & \textbf{Mean\hat{R}} & \textbf{Mean\hat{FM}} & \textbf{Mean\hat{S}} \\ \hline
% \textbf{BKP} & 82.14\% & 88.09\% & 80.83\% & 82.78\% & 80.83\% & 91.90\% \\ \hline
% \textbf{LBP} & \textbf{100\%} & \textbf{100\%} & \textbf{100\%} & \textbf{100\%} & \textbf{100\%} & \textbf{100\%} \\ \hline
% \textbf{HC} & 75\% & 83.33\% & 76.06\% & 73.56\% & 73.29\% & 87.89\% \\ \hline
% \textbf{MC} & 92.85\% & 95.23\% & 93.93\% & 93.93\% & 93.93\% & 96.49\% \\ \hline
% \end{tabular}
% }
% \end{center}
% \caption{Resultados generales }
% \label{tab:gral}
% \end{table}
% \vspace{-0.3in}
% \vspace{-0.1in}
Presentamos los resultados por clases sobre cada dataset en el Cuadro \ref{tab:classres} y realizamos un breve an\'alisis. 
Los resultados de la t\'ecnica BKP para cada clase muestran que para estos datos la clase 6 es la que m\'as le cost\'o identificar.
Los resultados de la t\'ecnica LBP para cada clase son muy alentadores ya que demuestran un excelente desempeño. 
Los resultados de la t\'ecnica HC para cada clase muestran que no pudo distinguirse muy acertadamente las clases 5 y 6.
Los resultados de la t\'ecnica MC para cada clase permiten ver que para este conjunto de datos se puede identificar de forma ideal la clase 0.

%Sin Specifficity
\begin{table}[htbp]
\begin{center}
% \centering
\resizebox{!}{70pt}{
\begin{tabular}{|c|c|c|c|c|c|c|}
\hline
\textbf{Feature Extraction} & \textbf{Clase} & \textbf{A} & \textbf{P} & \textbf{R} & \textbf{FM} \\ \hline
\textbf{BKP} & \textbf{0} & 92.85\% & 80\% & 100\% & 88.88\% \\ \hline
\textbf{BKP} & \textbf{5} & 89.28\% & 100\% & 76.92\% & 86.95\%  \\ \hline
\textbf{BKP} & \textbf{6} & 82.14\% & 62.5\% & 71.42\% & 66.66\%  \\ \hline
\textbf{LBP} & \textbf{0} & 100\% & 100\% & 100\% & 100\%  \\ \hline
\textbf{LBP} & \textbf{5} & 100\% & 100\% & 100\% & 100\%  \\ \hline
\textbf{LBP} & \textbf{6} & 100\% & 100\% & 100\% & 100\%  \\ \hline
\textbf{HC} & \textbf{0} & 92.85\% & 100\% & 81.81\% & 90\%  \\ \hline
\textbf{HC} & \textbf{5} & 78.57\% & 66.66\% & 50\% & 57.14\%  \\ \hline
\textbf{HC} & \textbf{6} & 78.57\% & 61.53\% & 88.88\% & 72.72\%  \\ \hline
\textbf{MC} & \textbf{0} & 100\% & 100\% & 100\% & 100\%  \\ \hline
\textbf{MC} & \textbf{5} & 92.85\% & 100\% & 81.81\% & 90\%  \\ \hline
\textbf{MC} & \textbf{6} & 92.85\% & 81.81\% & 100\% & 90\% \\ \hline
\end{tabular}
}
\end{center}
\caption{Resultados de cada t\'ecnica de extracci\'on de caracter\'isticas para cada clase}
\label{tab:classres}
\end{table}
\vspace{-0.1in}

  Algunas conclusiones que podemos extraer de los resultados son: 
  1) LBP es la t\'ecnica que presenta mejores resultados en este conjunto de datos ya que todas sus medidas son superiores al resto; lo que 
  nos dice que la informaci\'on local de textura tiene altas probabilidades de servir para caracterizar una imagen en base a su nivel de carga muscular;
  2) los segundos mejores resultados fueron obtenidos por la t\'ecnica MC, lo que nos indica que tambi\'en la informaci\'on de contorno sirve para 
   esta caracterizaci\'on;
  3) todos los resultados son superiores al 70\% y muchos de ellos mayores al 90\% lo que indica que las t\'ecnicas presentadas de extracci\'on de 
  caracter\'isticas en combinaci\'on con la t\'ecnica de clasificaci\'on elegidas son \'utiles para estimar de forma discreta en 
  3 valores el nivel de carga a partir de im\'agenes est\'aticas;
  4) tenemos la intuici\'on de que la causa del buen desempeño de LBP sea que la informaci\'on de textura caracteriza mejor variaciones en la piel.

%   %   \begin{itemize}
%   \squishlist
%    \item LBP es la t\'ecnica que presenta mejores resultados en este conjunto de datos ya que todas sus medidas son superiores al resto y perfectas: 
%    esto nos dice que  la informaci\'on local de textura tiene altas probabilidades de servir para caracterizar una imagen en base a su nivel de carga muscular.
%    \item Los segundos mejores resultados fueron obtenidos por la t\'ecnica MC: Esto nos indica que tambi\'en la informaci\'on de forma o contorno sirve para 
%    esta caracterizaci\'on
%    \item Todos los resultados son superiores al 70\% y muchos de ellos de ellos mayores al 90\%: Las t\'ecnicas presentadas de extracci\'on de caracter\'isticas 
%    en combinaci\'on con la t\'ecnica de clasificaci\'on elegidas son \'utiles para estimar de forma discreta en 3 valores el nivel de carga a partir de im\'agenes est\'aticas.
%   \squishend
% %   \end{itemize}

\vspace{-0.15in}
\section{Trabajos relacionados}\label{sec:trel}
\vspace{-0.15in}
Hasta donde los autores hemos podido relevar no se han encontrado trabajos en la literatura que resuelvan el problema de estimaci\'on de carga muscular mediante im\'agenes
de la piel, pero presentaremos en esta secci\'on los trabajos m\'as relacionados, principalmente en cuanto a la aplicaci\'on de Visi\'on Computacional para sensado muscular.

Encontramos en la literatura una serie de trabajos relacionados al sensado muscular a partir de im\'agenes de ultrasonido.
% \cite{peng20062d,chi2004ultrasound,barber2009validation,de2010individual,shi2010recognition,li2014estimation} 
Entre los problemas que resuelven los trabajos relevados podemos 
ver el de estimaci\'on de desplazamientos y movimientos musculares \cite{peng20062d,li2014estimation}; 
estimaci\'on de medidas internas del músculo 
(volumen, tamaño, di\'ametro, largo, \'area, fuerza de contracci\'on m\'axima voluntaria, entre otros) 
% m\'usculo 
% como son: volumen , tamaño, di\'ametro , largo , \'area, fuerza de contracci\'on m\'axima voluntaria, entre otros 
as\'i tambi\'en como 
la relaci\'on de estas medidas con los distintos niveles de actividad muscular 
\cite{chi2004ultrasound,barber2009validation,de2010individual} 
y finalmente identificar flexi\'on de los dedos y momento de ocurrencia (\cite{shi2010recognition}). 
Como se observa, los problemas que resuelven puntualmente estos trabajos estiman una gran cantidad de variables musculares, entre ellas medidas que
son correlativas a la carga muscular.
A pesar de que estos problemas son similares al que nosotros resolvemos, estos autores no atacan el problema puntual de estimar carga o esfuerzo muscular 
mediante im\'agenes de la piel. 
El uso de im\'agenes de ultrasonido no siempre es posible. Estas im\'agenes se deben capturar con dispositivos especiales, costosos y que requieren contacto directo con la 
piel imposibilitando el sensado a distancia.
Una ventaja de nuestro enfoque sobre estos trabajos es el uso de im\'agenes que si pueden ser capturadas a distancia y con una c\'amara convencional.
% A pesar de esto, las t\'ecnicas de visi\'on computacional aqu\'i utilizadas podr\'ian servir como gu\'ia y ser aplicadas para resolver
% nuestro problema en trabajos futuros.

Adem\'as, en la literatura es posible encontrar trabajos que realizan sensado muscular aplicando visi\'on computacional a im\'agenes externas como en 
\cite{zoccolan2001use,zoccolan2002using}. Estos trabajos resuelven el problema de estimar el nivel de contracci\'on muscular en sanguijuelas a partir de im\'agenes de 
movimiento microsc\'opicas de su piel. El problema resuelto por los autores mencionados esta relacionado con el que resolvemos en este trabajo, en cuanto a 
que tratan de cuantizar o caracterizar  la contracci\'on de un m\'usculo en base a im\'agenes externas. De todos modos si bien ellos resuelven un problema similar, no 
trabajan sobre piel humana y lo hacen a partir de videos utilizando t\'ecnicas que caracterizan el movimiento de los p\'ixeles realizando un seguimiento de un cuadro a otro
del video. 
Estos trabajos pueden ser un buen antecedente 
para caracterizar movimiento de texturas, y sus resultados podr\'ian ser aplicables a piel humana para caracterizar la contracci\'on muscular en personas 
y posteriormente
estimar el nivel de carga.

Como se puede observar, no se han encontrado trabajos en la literatura que resuelvan el problema puntual de estimar el nivel de carga muscular mediante im\'agenes de la piel. 
% A pesar de eso, s\'i hay trabajos en el \'area de EMG Prediction (usando sistemas de múltiples c\'amaras y con marcadores) 
% que es el pr\'oximo problema que queremos resolver, y 
% donde pretendemos proponer una soluci\'on  que sea mejor en t\'erminos de practicidad, costo e intrusividad
% % m\'as pr\'actica, econ\'omica y menos intrusiva de 
% al resolver el problema, usando informaci\'on visual (sin marcadores) que pueda ser obtenida con sensores econ\'omicos.

\vspace{-0.15in}
\section{Conclusiones y trabajo futuro}\label{sec:concl}
\vspace{-0.15in}
En este trabajo presentamos un enfoque que permite decidir con un alto grado de certeza qu\'e nivel de carga muscular de entre 3 posibles est\'a ejerciendo una persona a 
partir de una imagen de su brazo a piel descubierta. Esto fue posible realizando aprendizaje supervisado (SVM) 
sobre vectores de caracter\'isticas visuales de las im\'agenes obtenidos aplicando diversas t\'ecnicas para tal fin 
% (Bag of KeyPoints, Local Binary Patterns, Histogramas de Color, Momentos de Contorno). 
(BKP, LBP, HC, MC). 
Para el conjunto de
im\'agenes utilizadas en este trabajo, las t\'ecnicas de LBP y de MC dieron los mejores resultados. 
Sin embargo, todas las t\'ecnicas 
presentan una performance (Prec., Rec., F-M. y Acc.)
aceptable
 y superiores al 70\% (llegando al 100\% en el mejor caso). 
% Estimamos, basados en resultados preliminares (que no incluimos en este trabajo), que este enfoque
% podr\'a extender su resoluci\'on en la estimaci\'on hasta 5 niveles discretos de carga muscular manteniendo resultados relativamente buenos en las medidas de performance.
Cabe aclarar que el buen desempeño obtenido por todas las t\'ecnicas puede estar influenciado por la uniformidad en iluminaci\'on, escala, posici\'on y ventana de recuadro
impuesta a la captura de las im\'agenes de forma intencional con el objetivo de controlar lo mejor posible los datos de entrada. 
Cuando se implemente esta soluci\'on en entornos m\'as realistas,
puede ser necesario realizar normalizaciones en escala, iluminaci\'on y posici\'on a las im\'agenes o aplicar t\'ecnicas para extraer
caracter\'isticas que sean lo m\'as invariantes posible a cambios en estas propiedades.

% La soluci\'on de este problema es el punto de partida de una l\'inea de investigaci\'on que tiene como prop\'osito final la estimaci\'on autom\'atica y 
% a partir de informaci\'on 
% visual de variables de inter\'es biomec\'anicas.
Como trabajo futuro nos enfocaremos en mejorar la resoluci\'on del enfoque, realizando 
clasificaci\'on en 5 niveles de carga discreta, para posteriormente intentar realizar regresi\'on, 
estimando as\'i el peso real del objeto concreto que se est\'a levantando. A m\'as largo plazo se tiene como objetivo lograr predecir el nivel de la 
señal de EMG s\'olo a partir de informaci\'on visual,
% , para poder as\'i automatizar la medici\'on de esta variable de manera no intrusiva, 
ya que es de mucha utilidad en el estudio de la biomec\'anica.
% y que est\'a \'intimamente relacionada a las 
% fuerzas internas de los m\'usculos. 
Adem\'as mejoraremos el setting de experimentaci\'on trabajando sobre m\'as de un sujeto y realizando cross-validation para
mejorar la validez de los resultados y la significancia estad\'istica.
% (otra variable de inter\'es biomec\'anico).

\vspace{-0.1in}
\subsubsection{Agradecimientos}
\vspace{-0.1in}
El presente trabajo se llevo a cabo gracias a la financiaci\'on del autor Leandro Abraham a través de una beca de doctoral del CONICET 
bajo la direcci\'on de beca del Dr. Raymundo Forradellas (dedicaci\'on full time en Facultad de ingenier\'ia de UNCuyo),
en conjunto con la direcci\'on de tesis del Dr. Facundo Bromberg (dedicaci\'on full time en UTN-FRM y miembro de carrera de CONICET).
\vspace{-0.2in}

%The bibliography, done here without a bib file
%This is the old BibTeX style for use with llncs.cls
% \def\bibfont{\tiny}

% \bibliographystyle{splncs03}
%Alternative bibliography styles:
%the following does the same as above except with alphabetic sorting
%\bibliographystyle{splncs_srt}
%the following is the current LNCS BibTex with alphabetic sorting
%\bibliographystyle{splncs03}
%If you want to use a different BibTex style include [oribibl] in the document class line

% \begin{tiny}
% \begin{multicols}{2}
\bibliographystyle{splncs03}
\bibliography{bibliografia.bib}
% \end{multicols	}
% \end{tiny}

%  {\tiny
%  \bibliographystyle{abbrv}
%  \bibliography{bibliografia}
% }

\end{document}